\documentclass{article}

\usepackage[final, nonatbib]{neurips_2020}
\usepackage[utf8]{inputenc} 
\usepackage[T1]{fontenc}    
\usepackage{hyperref}       
\usepackage{url}            
\usepackage{booktabs}       
\usepackage{amsfonts}       
\usepackage{nicefrac}       
\usepackage{microtype}      

\usepackage{hyperref}
\hypersetup{
    colorlinks=true,
    linkcolor=blue,
    filecolor=magenta,      
    urlcolor=blue,
}

\usepackage{wrapfig}
\usepackage{graphicx}
\usepackage{caption}
\usepackage{subcaption}
\usepackage{floatrow}
\usepackage{booktabs}
\usepackage{varwidth}
\usepackage{todonotes}
\usepackage{caption}
\usepackage{subcaption}
\usepackage{wrapfig}

\usepackage[capitalize]{cleveref}

\newfloatcommand{capbtabbox}{table}[][\FBwidth]
\usepackage{blindtext}

\title{ChemBERTa-2: Towards Chemical Foundation Models}

\author{%
  Walid Ahmad\thanks{Equal contribution} \\
  Reverie Labs \\
  \texttt{walid@reverielabs.com} \\
  \And
  Elana Simon\footnotemark[1] \\
  Reverie Labs \\
  \texttt{elana@reverielabs.com} \\
  \And
  Seyone Chithrananda \\
  UC Berkeley  \\
  \texttt{seyonec@berkeley.edu} \\
  \And
  Gabriel Grand \\
  Reverie Labs \& MIT CSAIL \\
  \texttt{gg@mit.edu} \\
  \And
  Bharath Ramsundar \\
  Deep Forest Sciences \\
  \texttt{bharath@deepforestsci.com} \\
}

\begin{document}

\maketitle
\begin{abstract}
Large pretrained models such as GPT-3 have had tremendous impact on modern natural language processing by leveraging self-supervised learning to learn salient representations that can be used to readily finetune on a wide variety of downstream tasks \cite{bommasani2021opportunities}. We investigate the possibility of transferring such advances to \textit{molecular} machine learning by building a chemical foundation model, ChemBERTa-2, using the ``language" of SMILES. While labeled data for molecular prediction tasks is typically scarce, libraries of SMILES strings are readily available.

In this work, we build upon ChemBERTa \cite{chithrananda2020chemberta} by optimizing the pretraining process. We compare multi-task and self-supervised pretraining by varying hyperparameters and pretraining dataset size, up to 77M compounds from PubChem. To our knowledge, the 77M set constitutes one of the largest datasets used for molecular pretraining to date. We find that with these pretraining improvements, we are competitive with existing state-of-the-art architectures on the MoleculeNet \cite{wu2018moleculenet} benchmark suite. We analyze the degree to which improvements in pretraining translate to improvement on downstream tasks.
\end{abstract} 

\section{Motivation}



Over the past few years, transformers \cite{attention, bert} have emerged as popular architectures for learning self-supervised representations of molecules from text representations. ChemBERTa \cite{chithrananda2020chemberta} introduced a BERT-like transformer model that learns molecular fingerprints through semi-supervised pretraining and pretrained it on a dataset of 10M compounds. MolBERT \cite{fabian2020molecular} experiments with a number of different pretraining objectives on a dataset of 1.6M compounds. SMILES-BERT \cite{wang2019smiles} pretrains on 18.7M compounds from Zinc.

ChemBERTa-2 is a BERT-like transformer model \cite{roberta} that learns molecular fingerprints through semi-supervised pretraining of the language model. ChemBERTa-2 employs masked-language modelling (MLM) and multi-task regression (MTR) over a large corpus of 77 million SMILES strings, a well-known text representation of molecules. SMILES, is its own language, with a simple vocabulary, consisting of a series of characters representing atom and bond symbols, and very few grammar rules. \cite{smiles}. ChemBERTa-2 explores the scaling hypothesis that pretraining effectively on larger datasets can yield improved performance, using the largest training dataset in molecular representation learning. 


\section{Related Work}

While this paper and its preceding works explore transformer-based pretraining for molecular models, a parallel line of work has explored the used of graph-based pretraining methods. SNAP \cite{hu2019strategies} introduces graph pretraining methods based on node attribute masking and structural similarity. Grover \cite{rong2020self} scales graph-transformer pretraining to a 100 million parameter model pretrained on 10M compounds. MolGNet \cite{li2020learn} uses a message passing architecture to pretrain a 53 million parameter model on 11M compounds. 

A number of recent works have explored alternative pretraining methodologies including contrastive learning \cite{wang2022molecular}. Other work attempts to combine molecular graph and transformer based pretraining methodologies into a unified "dual" framework \cite{zhu2021dual}, or considers techniques inspired by neural machine translation by learning to translate between SMILES and InChi representations of a molecule \cite{winter2019learning}. Very recent work investigates whether large language models such as GPT-3, trained on non-chemical corpuses have learned meaningful chemistry \cite{white2022large}.



\section{Methods}

\begin{figure}
     \centering
     \begin{subfigure}[b]{0.45\textwidth}
         \centering
         \includegraphics[width=\textwidth]{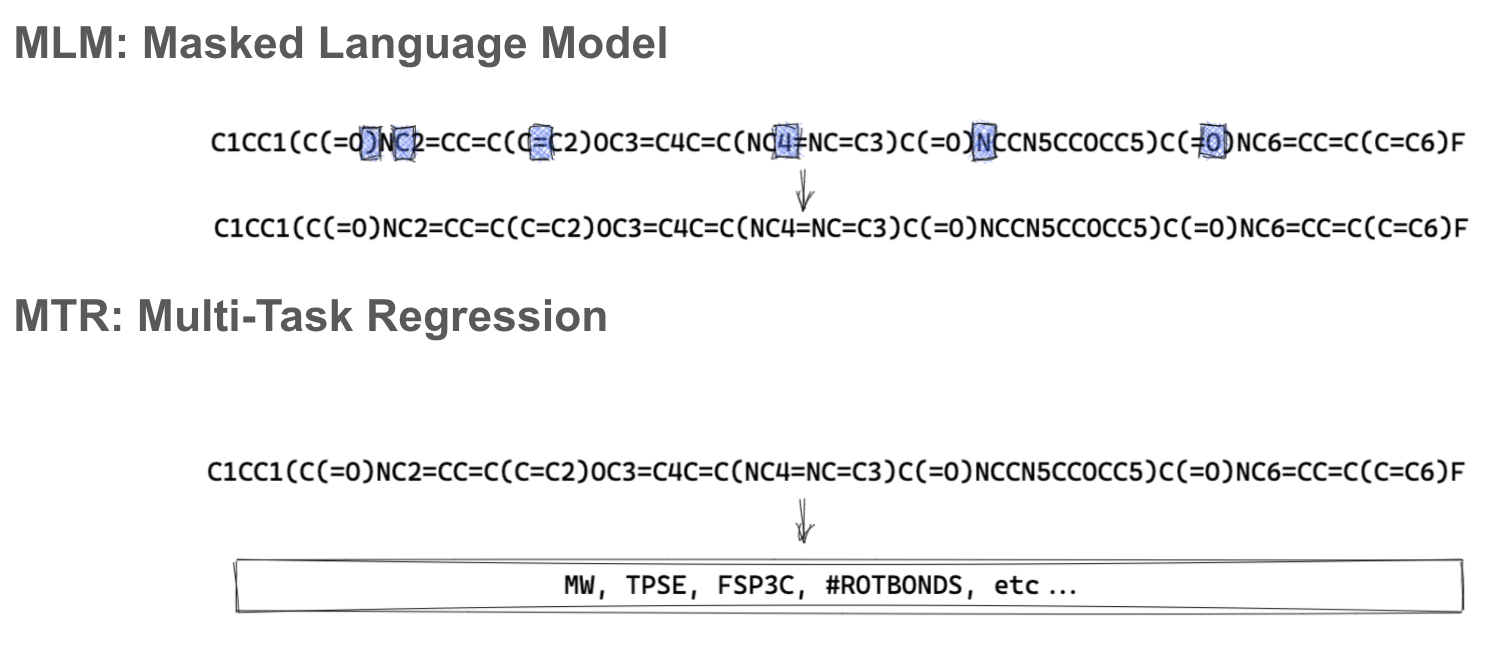}
         \caption{MLM vs. MTR}
         \label{fig:pretraining task diagram}
     \end{subfigure}
     \hfill
     \begin{subfigure}[b]{0.45\textwidth}
         \centering
         \includegraphics[width=\textwidth]{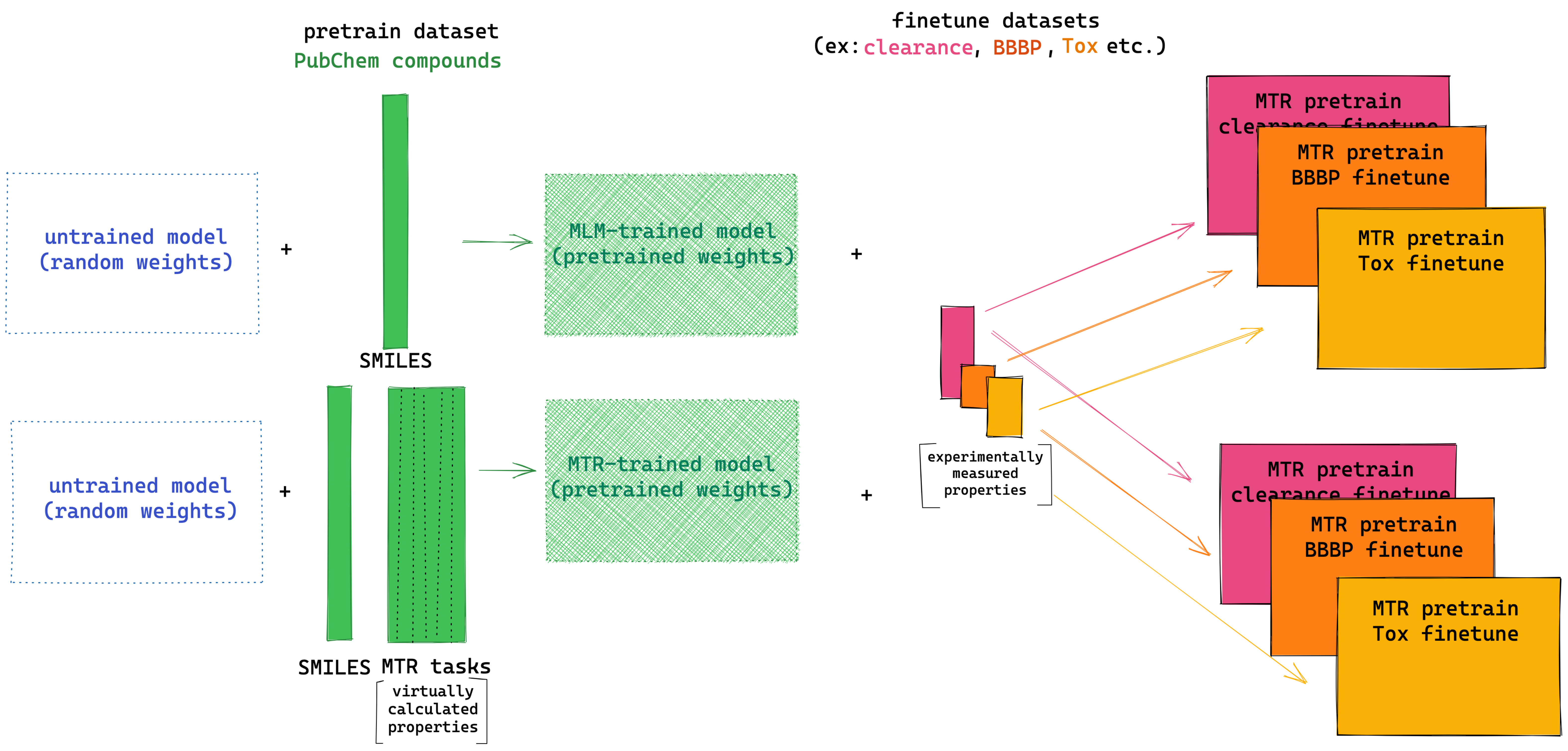}
         \caption{Training pipeline}
         \label{fig:training pipeline}
     \end{subfigure}
        \caption{a) An illustration of masked language modeling (MLM) and multitask regression (MTR) pretraining tasks. b) The training pipeline implemented to achieve results in this paper.}
        \label{fig:mlm and mtr}
\end{figure}




ChemBERTa-2 is based on the RoBERTa \cite{roberta} transformer implementation in HuggingFace \cite{huggingface}. We use the same training dataset of 77M unique SMILES from ChemBERTa \cite{chithrananda2020chemberta}. We canonicalize and globally shuffle the SMILES to facilitate large-scale pretraining. For validation, we first set aside a fixed set of 100k compounds. We divide the remaining dataset by sampling subsets of 5M, 10M and 77M (the full set), constituting three datasets to be used across both pretraining tasks.


\subsection{Pretraining Strategies and Setup}

\textbf{Masked Language Modeling:} We adopt the masked language modeling (MLM) pretraining procedure from RoBERTa, which masks 15\% of the tokens in each input string and trains the model to correctly identify them. We use a maximum vocab size of 591 tokens based on a dictionary of common SMILES characters and a maximum sequence length of 512 tokens.

\textbf{Multi-task Regression:} We compute a set of 200 molecular properties for each compound in our training dataset. These properties do not require any experimental measurements and can each be calculated from SMILES alone using RDKit \cite{rdkit}. We then train a multitask regression (MTR) architecture to predict these properties simultaneously. Because these tasks have very different scales and ranges, we mean-normalize the labels for each task prior to training.

\textbf{Pretraining Setup:} Models are trained on AWS EC2 instances equipped with Nvidia T4 GPUs. We set early stopping patience to equal one pass through the dataset, to ensure that for any dataset size, the model has an opportunity to see each compound at least once. 

\subsection{Hyperparameter Search}

Most language modeling architectures have hyperparameters that are tuned on datasets comprised of written and spoken language, such as English. SMILES on the other hand, have a very different grammatical structure. To ensure an adequate assessment of ChemBERTa-v2 performance, we conduct a thorough hyperparameter search (subject to compute constraints).

We select 50 random hyperparameter configurations, varying the hidden size, number of attention heads, dropout, intermediate size, number of hidden layers, and the learning rate. Models have between 5M and 46M parameters. Each configuration is trained on each of the MLM and MTR pretraining tasks, with the 5M dataset. Using the smallest dataset size ensures that we can train until convergence (as dictated by early stopping). From pretraining results, we select five configurations, with varying validation loss values, to train on the 10M and 77M sets. Five configurations are selected for MLM and MTR, independently from one another, with the objective of evaluating how pretraining loss can affect downstream performance. 

\subsection{Finetuning on MoleculeNet}

We evaluate our models on several regression and classification tasks from MoleculeNet \cite{wu2018moleculenet} selected to cover a range of dataset sizes (1.5K - 8.0K examples) and medicinal chemistry applications (brain penetrability, toxicity, solubility, and on-target inhibition). These included the BACE, Clearance, Delaney, Lipophilicity, BBBP, ClinTox, HIV, Delaney, and Tox21 datasets. For datasets with multiple tasks, we selected a single representative task: the clinical toxicity (CT\_TOX) task from ClinTox and the p53 stress-response pathway activation (SR-p53) task from Tox21. For each dataset, we generate an 80/10/10 train/valid/test split using the scaffold splitter from DeepChem \cite{deepchem}. We finetune models for up to 100 epochs with early stopping based on validation loss, and explore train-time hyperparameters via HuggingFace's built-in \texttt{optuna} optimization tooling, varying learning rate, random seed, and batch size. Finetuning task labels are normalized to have zero mean and unit standard deviation during training for regression tasks, and balanced class weights for classification tasks.

\section{Results}
In Table \ref{tab:main-results}, we share our results on MoleculeNet benchmarks. ChemBERTa-2 configurations are able to achieve competitive results on nearly all tasks, and outperform D-MPNN (\texttt{chemprop} implementation) on 6 out of 8 tasks.

\begin{table}[H]
\small
    \begin{tabular}{@{}lllll|llll@{}}
        \toprule
         & \multicolumn{1}{c}{\textbf{BACE}} & \multicolumn{1}{c}{\textbf{Clearance}} & \multicolumn{1}{c}{\textbf{Delaney}} & 
         \multicolumn{1}{c}{\textbf{Lipo}}
         & \multicolumn{1}{c}{\textbf{BACE}} & \multicolumn{1}{c}{\textbf{BBBP}} & \multicolumn{1}{c}{\textbf{ClinTox}} & 
         \multicolumn{1}{c}{\textbf{SR-p53}}  \\ 

        
        & \textit{RMSE} & \textit{RMSE} & \textit{RMSE} & \textit{RMSE} & \textit{ROC} 
        & \textit{ROC} & \textit{ROC} & \textit{ROC} \\
         \midrule
        D-MPNN & 2.253 & 49.754 & 1.105 & 1.212 & 0.812 & 0.697 & \textbf{0.906} & 0.719\\
        RF & \textbf{1.3178} & 52.0770 & 1.7406 & 0.9621 & \textbf{0.8507} & 0.7194 & 0.7829 & 0.724 \\
        GCN & 1.6450 & 51.2271 & 0.8851 & 0.7806 & 0.818 & 0.676 & 0.907 & 0.688 \\
        ChemBERTa-1 & & & & & & 0.643 & 0.733 & 0.728 \\
        \midrule

        \textbf{ChemBERTa-2}\\
        MLM-5M & 1.451 & 54.601 & 0.946 & 0.986 & 0.793 & 0.701 & 0.341 & 0.762\\
        MLM-10M & 1.611 & 53.859 & 0.961 & 1.009 & 0.729 & 0.696 & 0.349 & 0.748\\
        MLM-77M & 1.509 & 52.754 & 1.025 & 0.987 & 0.735 & 0.698 & 0.239 & 0.749\\
        MTR-5M & 1.477 & 50.154 & 0.874 & 0.758 & 0.734 & \textbf{0.742} & 0.552 & \textbf{0.834}\\
        MTR-10M & 1.417 & 48.934 & \textbf{0.858} & \textbf{0.744} & 0.783 & 0.733 & 0.601 & 0.827\\
        MTR-77M & 1.363 & \textbf{48.515} & 0.889 & 0.798 & 0.799 & 0.728 & 0.563 & 0.817\\
        \bottomrule

        
    \end{tabular}

\caption{\label{tab:main-results}Comparison of ChemBERTa-2 pretrained on different tasks (MLM and MTR) and on different dataset sizes (5M, 10M, and 77M), vs. existing architectures on selected MoleculeNet tasks. We report ROC-AUC  ($\uparrow$) for classification and RMSE ($\downarrow$) for regression tasks. D-MPNNs were trained with the \texttt{chemprop} \cite{yang2019analyzing} library. We could not benchmark easily against Grover \cite{rong2020self} due to differences in benchmarking procedures.}
\end{table}

\subsection{Selection of Pretraining Method}
\begin{figure}[h]
\caption{Comparing MLM and MTR pretrain losses}
\centering
\includegraphics[width=0.5\textwidth]{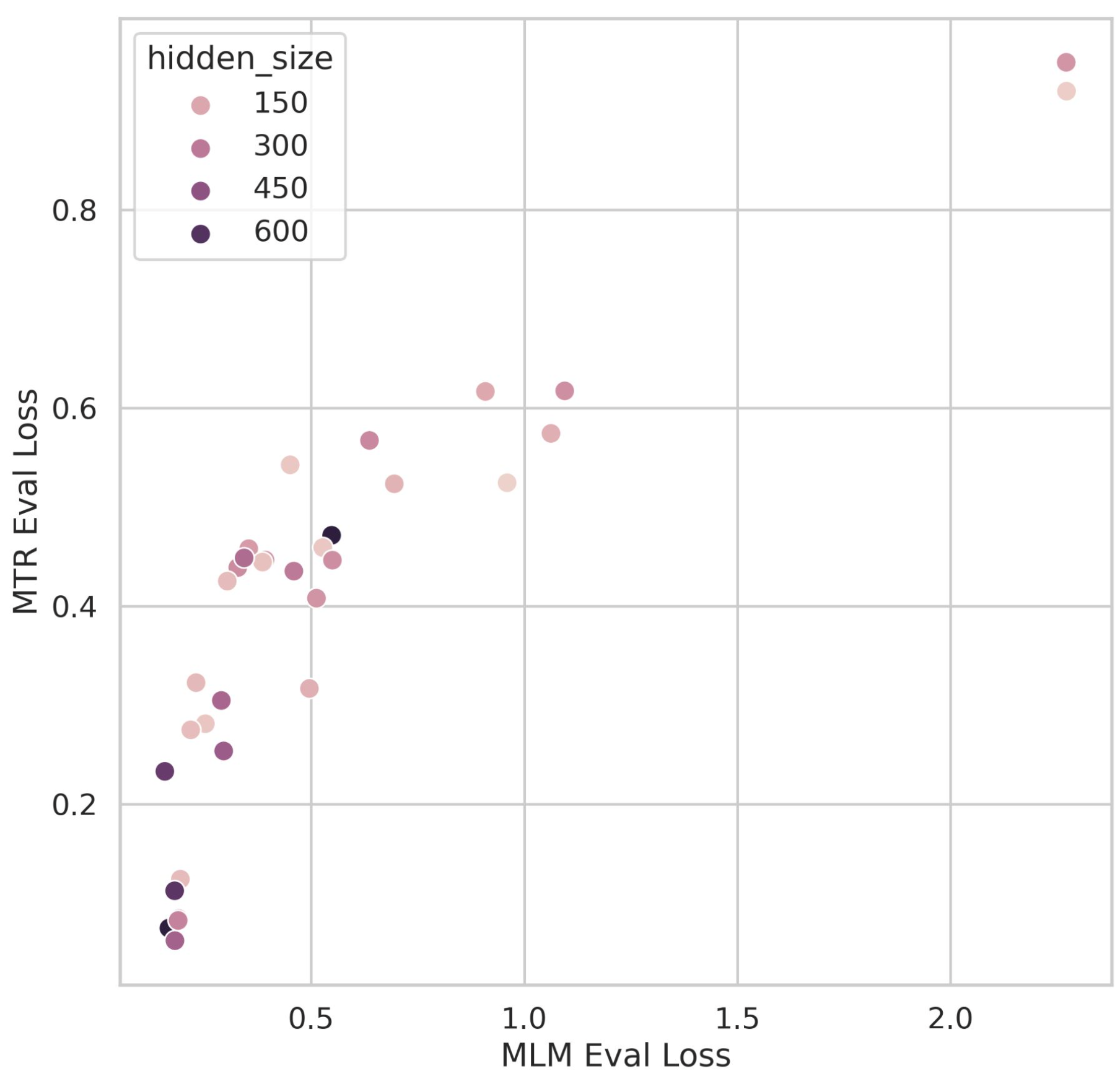}
\label{fig:mlm vs mtr pretrain loss}
\end{figure}

On every downstream finetuning task, models pretrained on the MTR task tend to perform better than models pretrained on the MLM task. However, in our current implementation, MTR training is substantially slower than MLM, due to the increased dataset size from the 200-element label vector. To address this, we observe that MLM pretraining loss corresponds very well with MTR pretraining loss for a given architecture. In Figure \ref{fig:mlm vs mtr pretrain loss}, we can see MTR vs MLM loss for a given configuration on the 5M dataset (where the same configurations were trained for both tasks). Thus, an architecture search can first be done via MLM pretraining, and the selected architecture(s) can then be trained on the MTR task for superior downstream performance.

\begin{figure}
     \centering
     \begin{subfigure}[b]{0.4\textwidth}
         \centering
         \includegraphics[width=\textwidth]{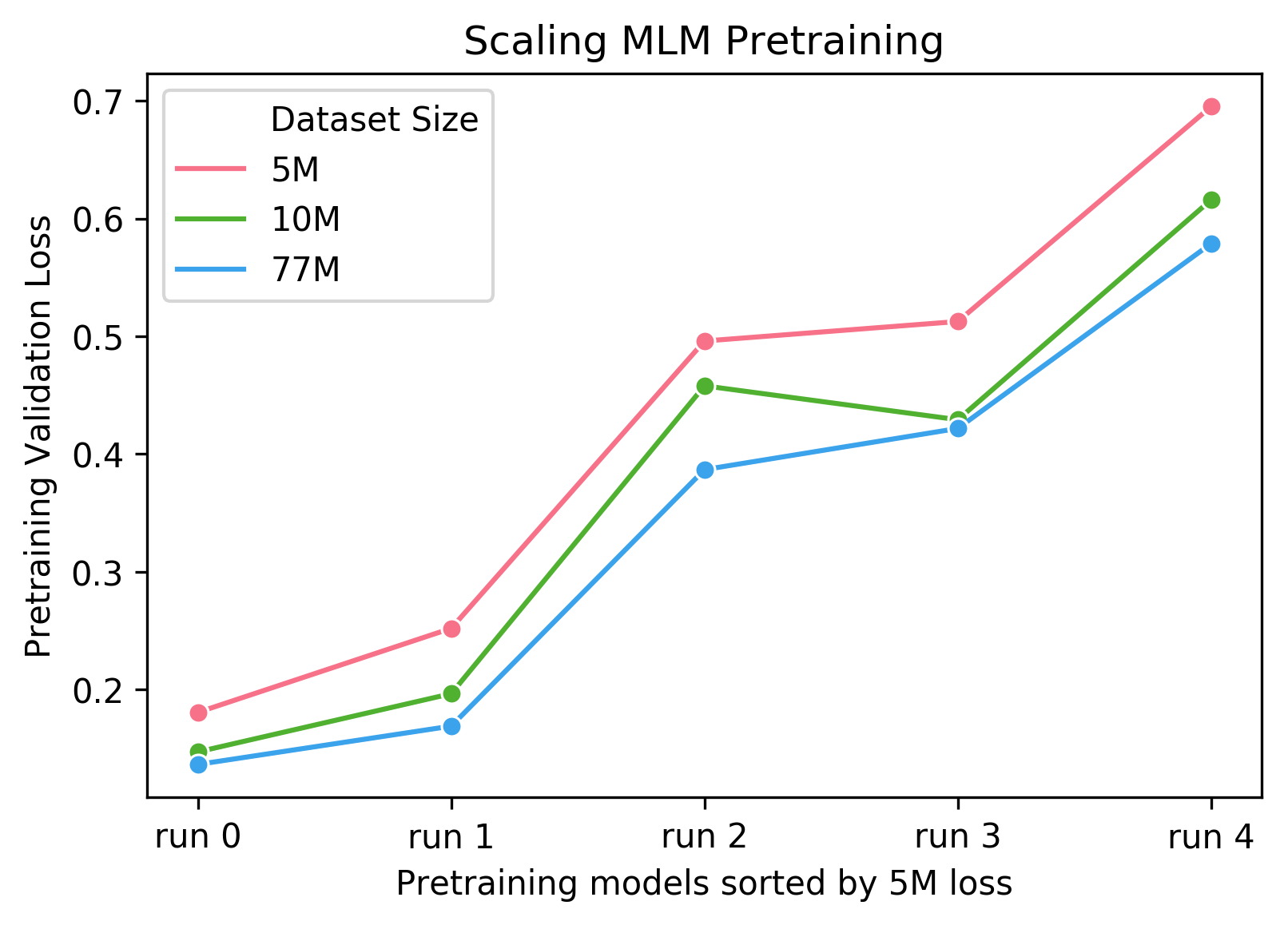}
         \label{fig:scaling mlm}
     \end{subfigure}
     \begin{subfigure}[b]{0.4\textwidth}
         \centering
         \includegraphics[width=\textwidth]{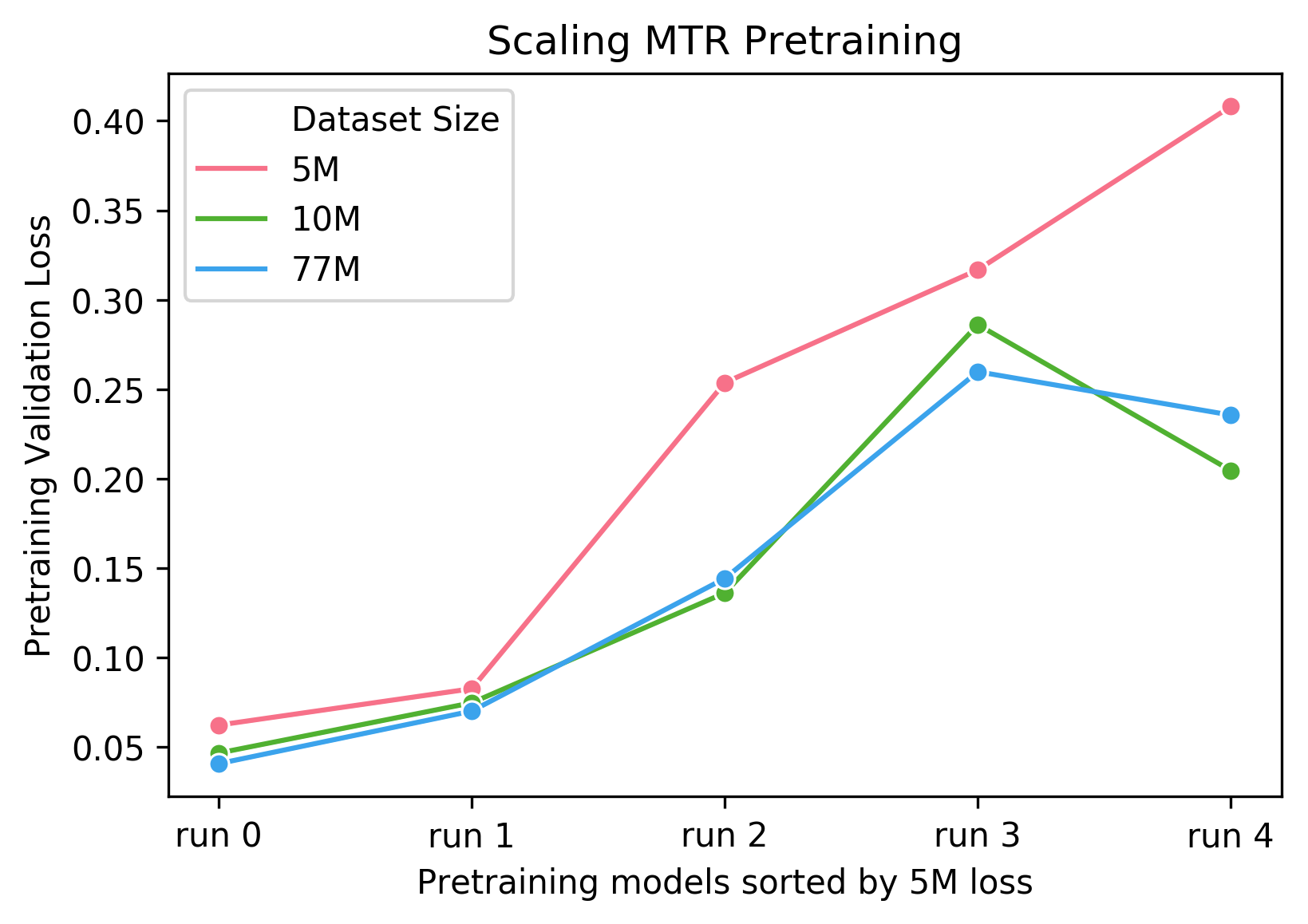}
         \label{fig:scaling mtr}
     \end{subfigure}
        \caption{Pretrain losses for each of the 5 model configurations that were trained on all three datasets (5M, 10M, and 77M). MLM configurations are on the left, and MTR on the right. The configurations are sorted by their loss when training with 5M compounds along the x-axis. Note that there is considerable performance variability across runs.}
        \label{fig:scaling mlm and mtr}
\end{figure}

In our experiments, we observe consistent improvements to pretraining loss with increased dataset size. As we see in Figure \ref{fig:scaling mlm and mtr}, training a model until convergence on 77M unique smiles instead of 5M can improve the pretraining loss by 25-35\%, an observation which holds across models with varying levels of performance for both MLM and MTR pretraining. 

\begin{figure}
     \centering
     \begin{subfigure}[b]{0.33\textwidth}
         \centering
         \includegraphics[width=\textwidth]{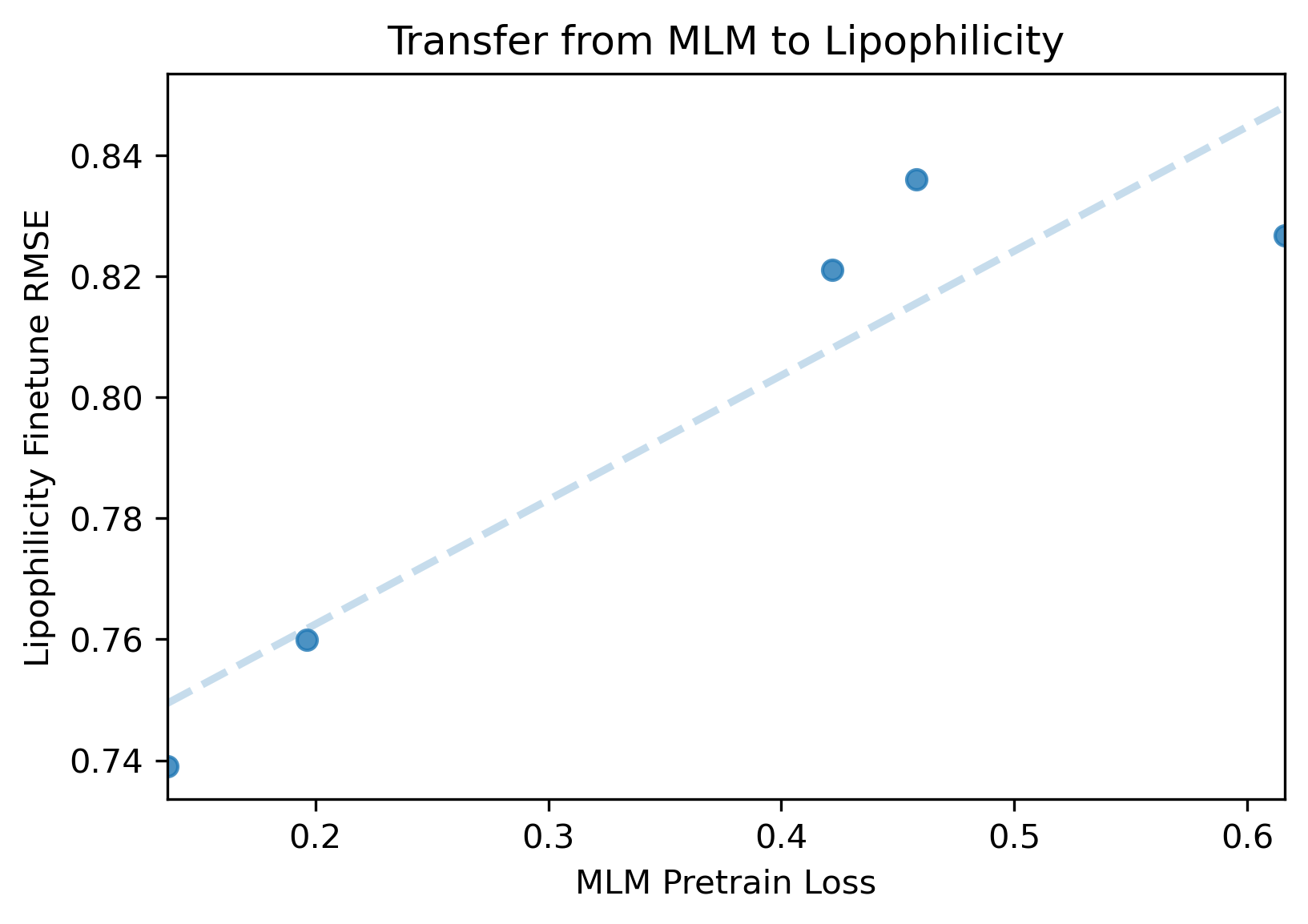}
         \label{fig:lipo mlm}
     \end{subfigure}
     \begin{subfigure}[b]{0.33\textwidth}
         \centering
         \includegraphics[width=\textwidth]{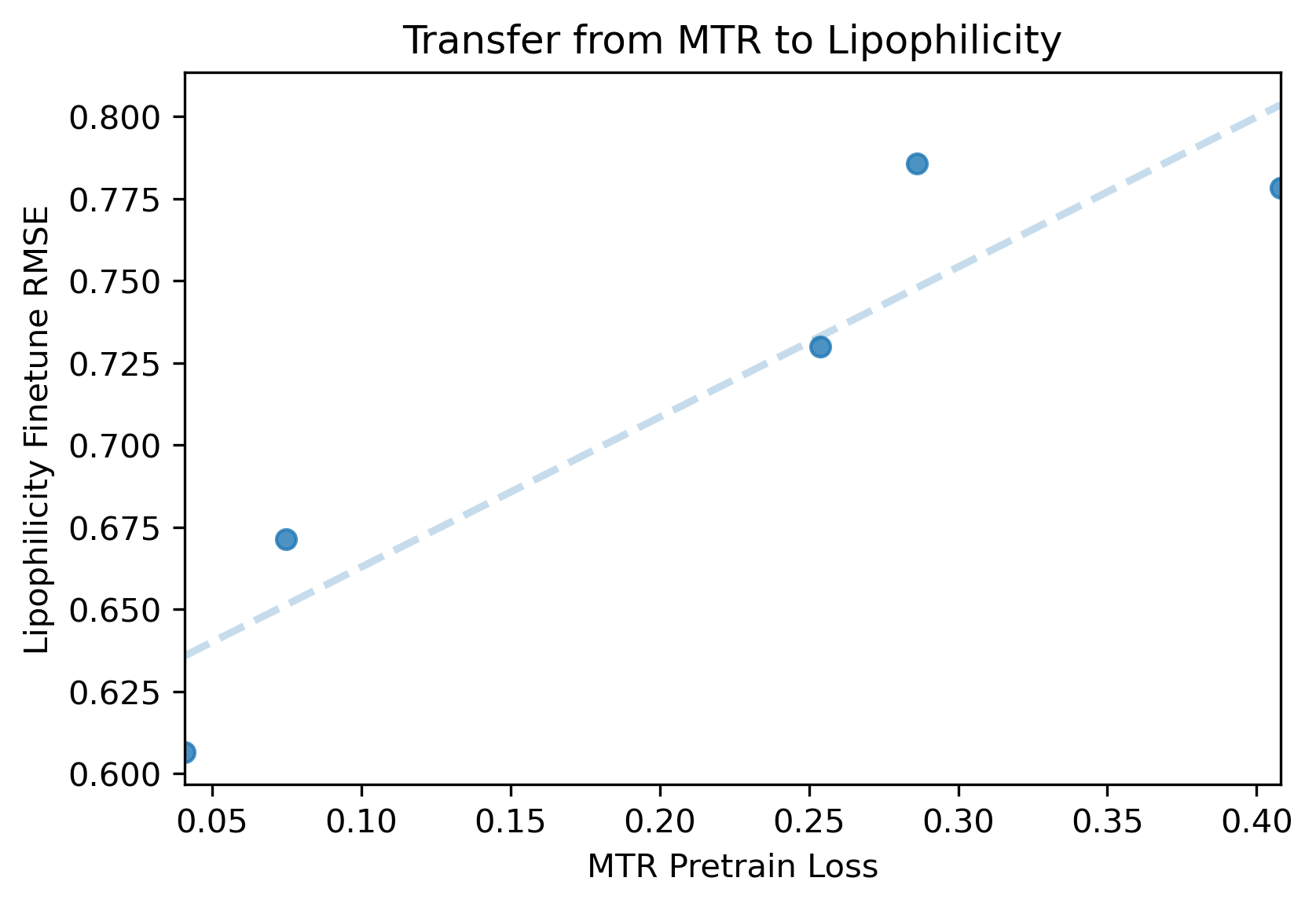}
         \label{fig:lipo mtr}
     \end{subfigure}
     \begin{subfigure}[b]{0.33\textwidth}
         \centering
         \includegraphics[width=\textwidth]{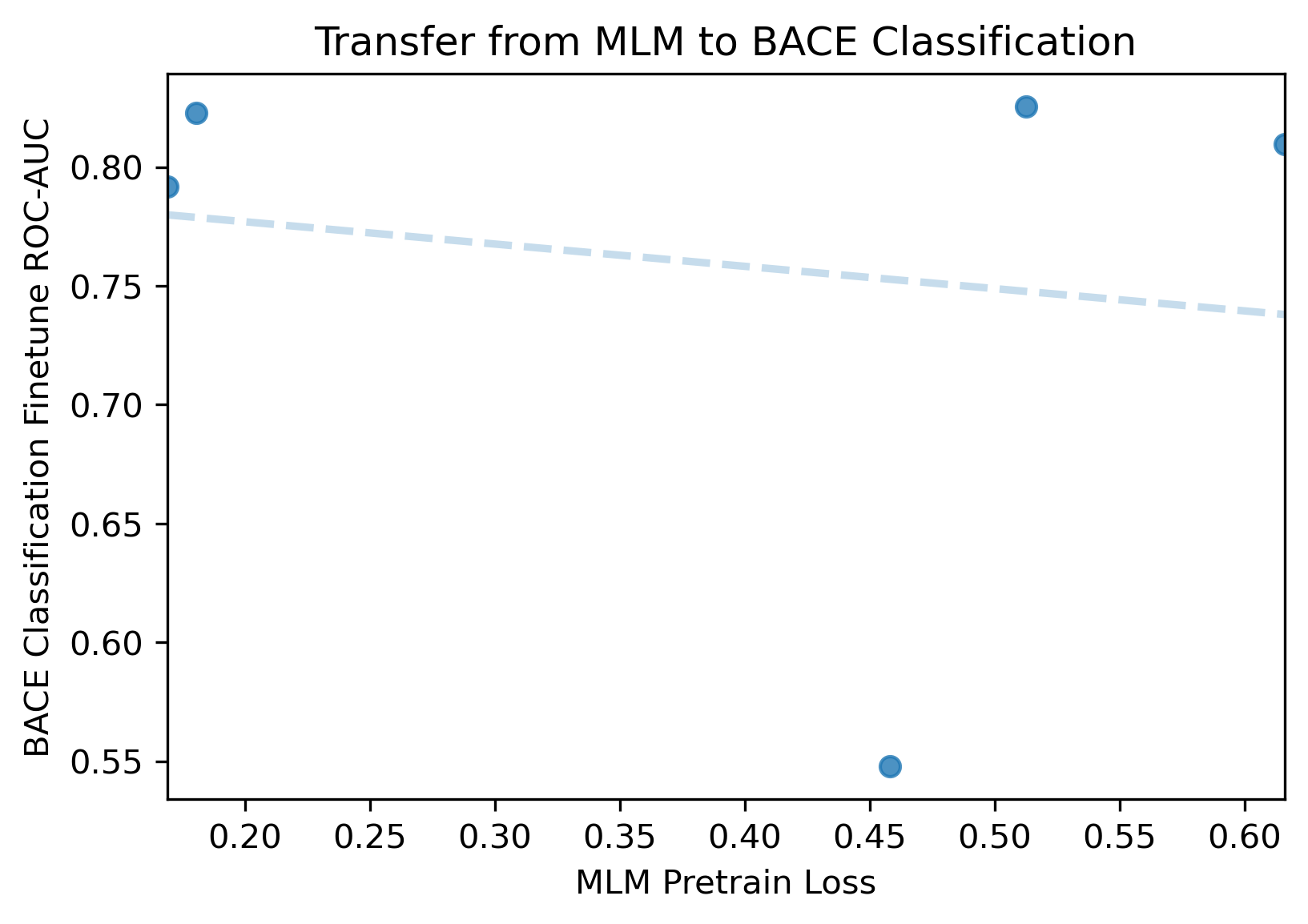}
         \label{fig:bace mlm}
     \end{subfigure}
     \begin{subfigure}[b]{0.33\textwidth}
         \centering
         \includegraphics[width=\textwidth]{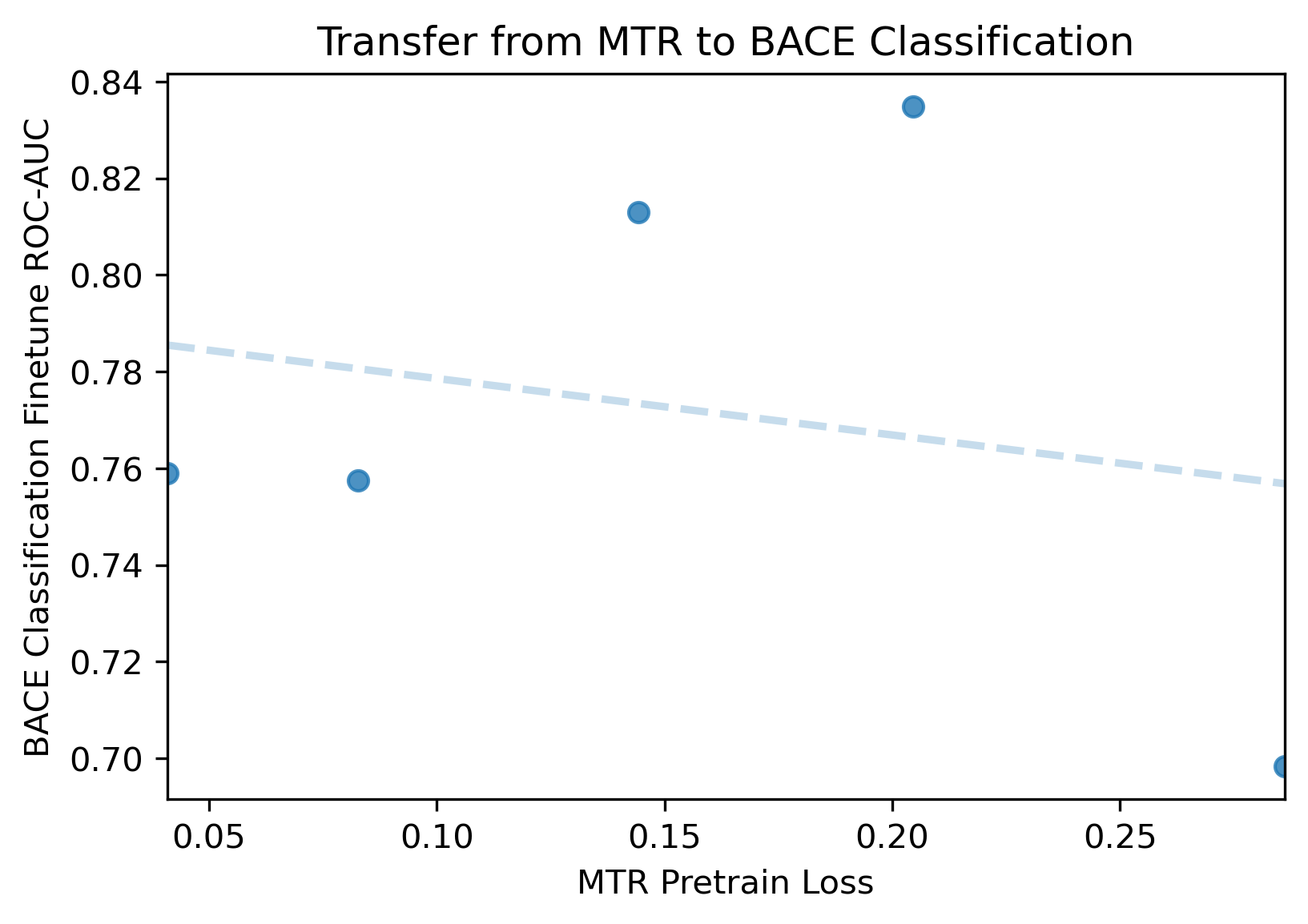}
         \label{fig:bace mtr}
     \end{subfigure}
        \caption{Finetuning performance versus pretraining loss. Left Column: MLM Pretraining, Right Column: MTR Pretraining. Top Row: Lipophilicity Finetune, RMSE ($\downarrow$), Bottom Row: BACE Classification Finetune, ROC-AUC ($\uparrow$). The dotted lines represent linear models fit to the datapoints.}
        \label{fig:lipo and bace pt to ft}
\end{figure}

We find that the degree to which improving performance on the pretraining tasks transfers to downstream tasks, varies by dataset. In Figure \ref{fig:lipo and bace pt to ft} we show two examples of transfer learning from pretrain tasks to finetune tasks with varying degrees of success. Improving (decreasing) pretraining loss for MLM and MTR leads to almost linear improvements (decrease) in Lipophilicity RMSE. This pattern does not hold as clearly for BACE Classification.


\section{Dimension Reduction of ChemBERTa Embeddings}

We used UMAP \cite{umap} to inspect the representations learned by pre-trained ChemBERTa models on the BACE and BBBP tasks, and contrast them to ECFP embeddings. We aim to see how well pre-trained language models without any further fine-tuning on MolNet benchmarks perform at clustering embeddings according to their labels. We drop large extra fragments in SMILES (using RDKit's LargeFragmentChooser) to avoid presence of salts, which are irrelevant to blood-brain barrier permeability, before generating both transformer and ECFP embeddings. 

We parameterize a UMAP model based on Jaccard distance with the following settings \texttt{metric = "jaccard", n\char`_neighbors = 25, n\char`_components = 2, low\char`_memory = False, min\char`_dist = 0.001}. We find that on average, ChemBERTa embeddings from both pre-trained masked-language and multi-task regression models are a stronger prior representation for a variety of downstream tasks to be fine-tuned on.

\begin{figure}[h!]
  \begin{subfigure}[b]{0.32\textwidth}
    \includegraphics[width=0.9\linewidth]{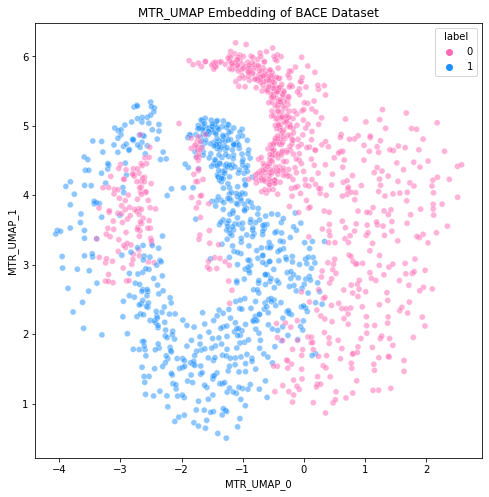}
    \caption{ }
    \label{fig:1}
  \end{subfigure}
  \begin{subfigure}[b]{0.32\textwidth}
    \includegraphics[width=0.9\linewidth]{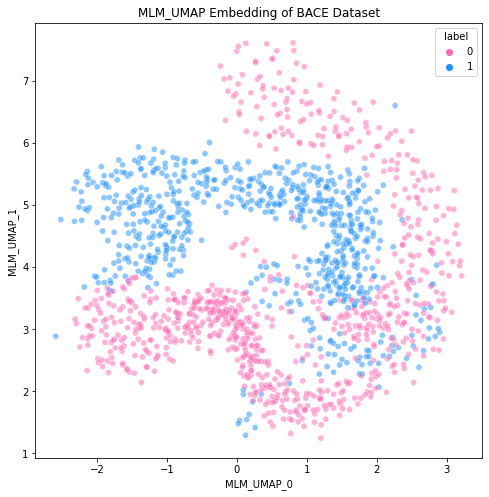}
    \caption{ }
    \label{fig:2}
  \end{subfigure}
  \begin{subfigure}[b]{0.32\textwidth}
    \includegraphics[width=0.9\linewidth]{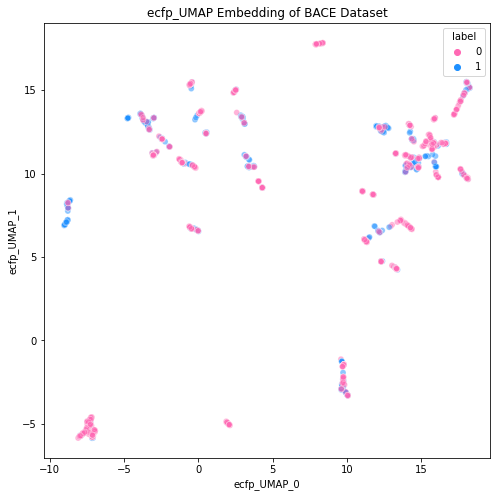}
    \caption{ }
    \label{fig:3}
  \end{subfigure}
  \begin{subfigure}[b]{0.32\textwidth}
    \includegraphics[width=0.9\linewidth]{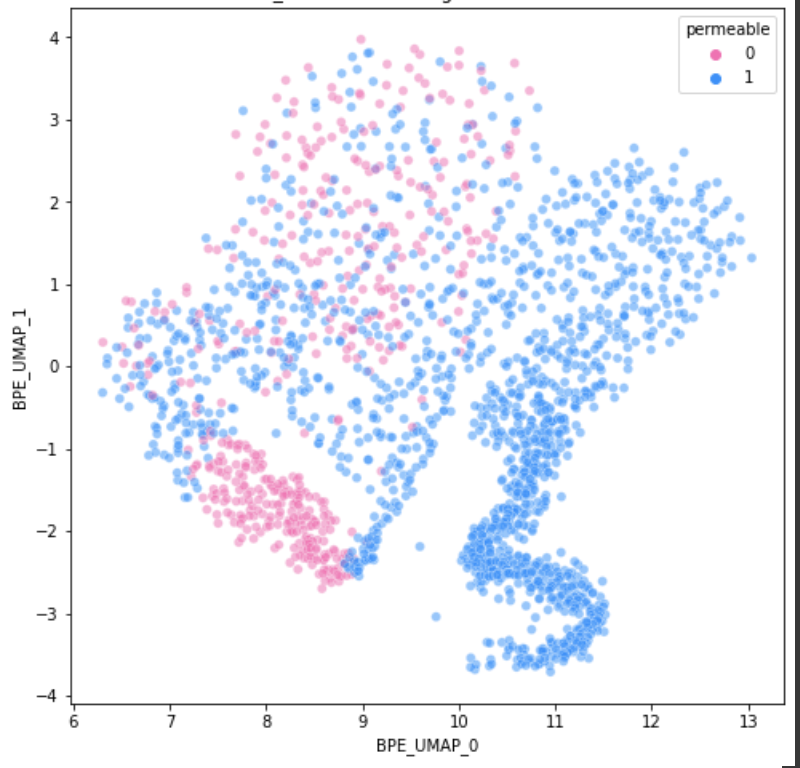}
    \caption{ }
    \label{fig:4}
  \end{subfigure}
  \begin{subfigure}[b]{0.32\textwidth}
    \includegraphics[width=0.9\linewidth]{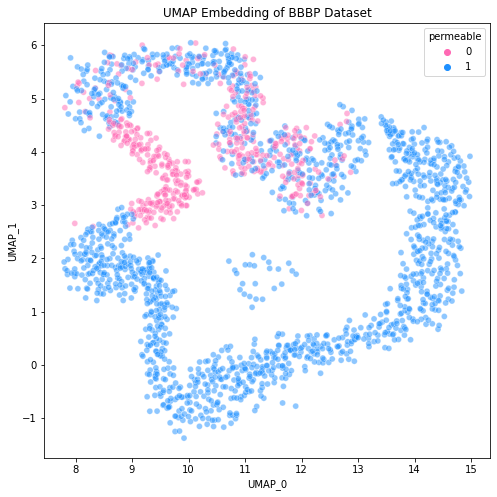}
    \caption{ }
    \label{fig:5}
  \end{subfigure}  
  \begin{subfigure}[b]{0.32\textwidth}
    \includegraphics[width=0.9\linewidth]{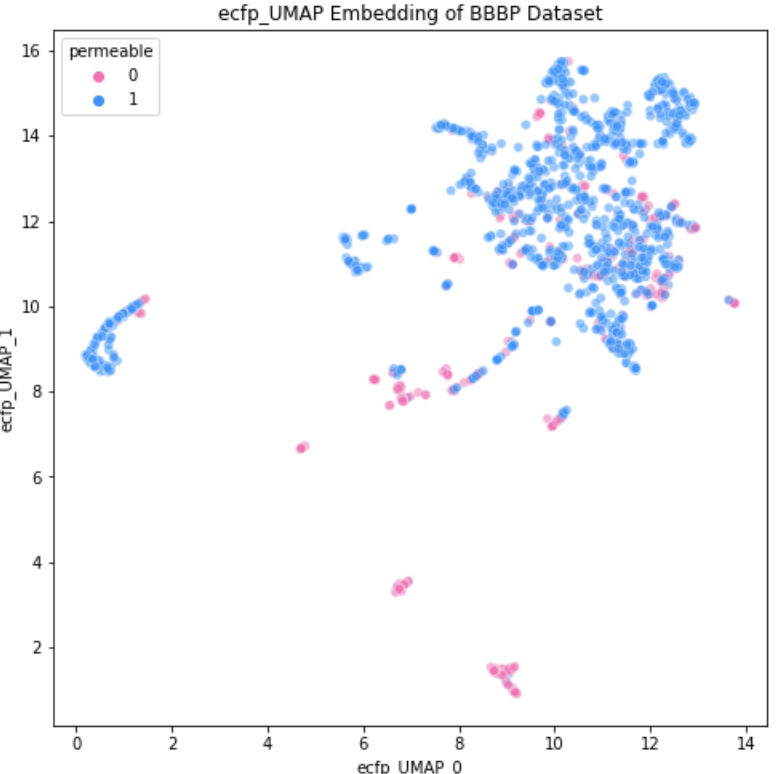}
    \caption{ }
    \label{fig:6}
  \end{subfigure}

  \caption{(a) MTR-77M embeddings fit using UMAP on BACE classification task. (b) MLM-77M embeddings fit using UMAP on BACE classification task. (c) ECFP embeddings fit using UMAP on BACE classification task. (d) MTR-77M embeddings fit using UMAP on BBBP classification task. (e) MLM-77M embeddings fit using UMAP on BBBP classification task (f) ECFP embeddings fit using UMAP on BBBP classification task}
\end{figure}

\section{Discussion}
In this work, we introduce ChemBERTa-2, an updated transformer architecture for molecular property prediction. By more deeply exploring the pretraining pipeline, we are able to achieve more competitive baseline results on downstream tasks, while extracting insights into pretraining strategies for language models. We use the more efficient MLM pretraining to select suitable hyperparameters for MTR pretraining and investigate the relationship between pretraining loss and downstream performance. Future work will benchmark against Grover and other graph based architectures and extend pretraining to larger datasets.

We also observe that certain finetuning tasks benefited greatly from pretraining improvements (either due to increased training time, increased pretraining dataset size, or varied hyperparameters) whereas others did not. The varied degree of transfer could depend on the type of task modeled (ex: solubility vs toxicity), the structural features of molecules in each dataset, the size of each dataset, or potentially even other features of the dataset we have not considered. We do not explore this phenomena in this work but note that this is an important area for future exploration. We have mostly focused on ways to improve molecular transfer learning by optimizing model pretraining procedures, but just as critical is to understand the conditions under which datasets might meaningfully benefit from pretraining.

We open source the trained models from this project. While recent work has highlighted the dual use risk \cite{urbina2022dual} of chemical models, for now we believe that the challenge of synthesizing a novel molecule limits the potential harms from releasing updated models. The balance may shift in future continuation of this research, and we will continue to assess the risks of dual use for future open source releases.

We also note that the terminology of foundation model has drawn criticism due to the fact that the data large language models are trained on is typically heavily biased. We note that the data we pretrain on is drawn from more fundamental chemical calculations, so we feel that the terminology "chemical foundation model" is appropriate even if the use of the term "foundation model" may be less appropriate for models such as GPT-3.





\bibliography{main}
\bibliographystyle{unsrt}

\section{Appendix}
\subsection{Training Tips}
These are some tips and things that we found useful when running large scale pretraining experiments across different machines. 
1) Due to availability of machines we train a few models on smaller machines that require decreasing the batch size to fit into memory. When doing this we decrease the learning rate accordingly to mitigate the effects of using different batch sizes.
2) When training MTR, our input data starts in CSV format. We find that using the default HuggingFace CSV data loader is notably slower than their text-based data loader so we develop a custom wrapper around their text-based loader that parses the text into tabular format. For large-scale pretraining, this makes a big difference for efficiency.
3) To save on cost we use AWS spot instances which sometimes get interrupted. HuggingFace has a nice system for re-starting models part-way through training. Using this is critical for training models that take a long time and get interrupted periodically.

\end{document}